\documentclass[runningheads]{llncs}
\usepackage{graphicx}

\usepackage{tikz}
\usepackage{comment} 
\usepackage{amsmath,amssymb} 
\usepackage{color}

\begin{document}
\pagestyle{headings}
\mainmatter
\def\ECCVSubNumber{2209}  

\title{MimicDet: Bridging the Gap Between One-Stage and Two-Stage Object Detection} 

\titlerunning{MimicDet}
%
\author{Xin Lu\inst{1} \and
Quanquan Li\inst{1} \and
Buyu Li\inst{2} \and
Junjie Yan\inst{1}}
\authorrunning{X. Lu et al.}
%
\institute{SenseTime Research
\and
The Chinese University of Hong Kong\\
\email{\{luxin,liquanquan,yanjunjie\}@sensetime.com} \quad \email{byli@ee.cuhk.edu.hk}}
\maketitle

\begin{abstract}

Modern object detection methods can be divided into one-stage approaches and two-stage ones. One-stage detectors are more efficient owing to straightforward architectures, but the two-stage detectors still take the lead in accuracy. Although recent work try to improve the one-stage detectors by imitating the structural design of the two-stage ones, the accuracy gap is still significant. In this paper, we propose MimicDet, a novel and efficient framework to train a one-stage detector by directly mimic the two-stage features, aiming to bridge the accuracy gap between one-stage and two-stage detectors. Unlike conventional mimic methods, MimicDet has a shared backbone for one-stage and two-stage detectors, then it branches into two heads which are well designed to have compatible features for mimicking. Thus MimicDet can be end-to-end trained without the pre-train of the teacher network. And the cost does not increase much, which makes it practical to adopt large networks as backbones. We also make several specialized designs such as dual-path mimicking and staggered feature pyramid to facilitate the mimicking process.
Experiments on the challenging COCO detection benchmark demonstrate the effectiveness of MimicDet. It achieves 46.1 mAP with ResNeXt-101 backbone on the COCO test-dev set, which significantly surpasses current state-of-the-art methods.

\keywords{Object Detection, Knowledge Distillation}
\end{abstract}

\section{Introduction}

\label{introduction}

In recent years, the computer vision community has witnessed significant progress in object detection with the development of deep convolutional neural networks (CNN). Current state-of-the-art detectors~\cite{girshick2015fast,ren2015faster,lin2017feature,he2017mask,cai2018cascade,zhang2018single,lin2017focal} show high performance on several very challenging benchmarks such as COCO~\cite{lin2014microsoft} and Open Images~\cite{kuznetsova2018open}. These detectors can be divided into two categories, one-stage detectors and two-stage ones. 
Modern one-stage detectors~\cite{lin2017focal,zhang2018single,tian2019fcos,zhu2019feature,chen2019revisiting} usually adopt a straightforward fully convolutional architecture, and the outputs of the network are classification probabilities and box offsets(w.r.t. pre-defined anchor box) at each spatial position.
While two-stage detectors have a more complicated pipeline~\cite{ren2015faster,lin2017feature,he2017mask,lu2019grid}. It first filters out the regions that have high probability to contain an object from the entire image with region proposal network (RPN)~\cite{ren2015faster}. Then the proposals are fed into the region convolutional network (R-CNN)~\cite{girshick2015fast} and get their classification score and the spatial offsets.
One-stage detectors are more efficient and elegant in design, but currently the two-stage detectors have domination in accuracy.

Compared to one-stage detectors, the two-stage ones have the following advantages: 1) By sampling a sparse set of region proposals, two-stage detectors filter out most of the negative proposals; while one-stage detectors directly face all the regions on the image and have a problem of class imbalance if no specialized design is introduced. 2) Since two-stage detectors only process a small number of proposals, the head of the network (for proposal classification and regression) can be larger than one-stage detectors, so that richer features will be extracted. 3) Two-stage detectors have high-quality features of sampled proposals by use of the RoIAlign~\cite{he2017mask} operation that extracts the location consistent feature of each proposal; but different region proposals can share the same feature in one-stage detectors and the coarse and spatially implicit representation of proposals may cause severe feature misalignment. 4) Two-stage detectors regress the object location twice (once on each stage) and the bounding boxes are better refined than one-stage methods.

We note that recent works try to overcome the drawbacks of one-stage detectors by imitating the structural design of the two-stage ones. For example, RefineDet~\cite{zhang2018single} tries to imitate the two-stage detectors by introducing cascade detection flow into a one-stage detector. Based on RefineDet, AlignDet~\cite{chen2019revisiting} proposes the RoIConv layer to solve the feature misalignment problem. The RoIConv layer imitates the RoIAlign~\cite{he2017mask} operation to match the feature and its corresponding anchor box with accurate and explicit location mapping. Although these works alleviate part of the disadvantages of traditional one-stage detectors, there still leaves a big gap to the two-stage detectors.

To further close the performance gap between one-stage and two-stage detectors, one natural idea is to imitate the two-stage approach not only in structure design, but also in feature level. An intuitive approach is to introduce the knowledge distillation(a.k.a mimic) workflow. However, unlike common mimic tasks, the teacher and student models are heterogeneous here. The features extracted from the backbone in teacher and student networks have different representing areas and also a big semantic gap. Thus it is inappropriate to directly apply existing mimic method in this heterogeneous situation. Moreover, conventional mimic workflow in object detection has many problems in practice. It needs to pre-train a teacher model with large backbone and then take the features from the teacher model as a supervision during the training of the student model. The entire workflow is complicated and the training efficiency is much lower. Besides, state-of-the-art object detectors usually adopt a relatively high resolution for input images, therefore forwarding both teacher and student networks simultaneously is hard in practice due to the large computation and GPU memory cost. In addition, traditional knowledge distillation methods usually use a powerful model as teacher to improve the performance of a student with small model, so it suffers scale-up limiation when the student model is comparatively large.

In this paper, we propose a novel training framework, named MimicDet, which can efficiently and significantly close the accuracy gap between the one-stage and two-stage detectors. Specifically, the network contains both one-stage detection head and two-stage detection head during training. The two-stage detection head, called T-head (teacher head), is a network branch with large amount of parameters, in order to extract high-quality features for sparsely sampled proposal boxes. The one-stage detection head, called S-head (student head), light-weighted branch for detecting all dense anchor boxes. Since the original teacher and student networks are heterogeneous, their feature maps have a spatial misalignment. To handle this problem, we exploit the guided deformable conv~\cite{wang2019region} layer where each convolution kernel can have an offset that computed by a lightweight neural network. Different from the AlignDet~\cite{chen2019revisiting}, the deformable offset can be optimized by mimicking the location consistent features generated by T-head. With the features matched, the similarity loss of the feature pairs will be optimized together with detection losses in MimicDet. Thus the features of S-head can acquire better properties from features of T-head. 
During inference, the T-head is discarded, which means a pure one-stage detector is implemented for object detection. This mechanism ensures that MimicDet inherits both high-efficiency and high-accuracy from the two architectures.
Different from the conventional mimic methods in object detection~\cite{romero2014fitnets,li2017mimicking}, teacher and student share the same backbone in MimicDet and the mimic is between different detection heads instead of different backbones. Thus it does not need to pre-train a teacher model or require a stronger backbone model to serve as the teacher. These properties make MimicDet much more efficient and can be extended to larger models.

To further improve the performance of MimicDet, several specialized designs are also proposed. We design decomposed detection heads and conduct mimicking in classification and regression branches individually. The decomposed \textit{dual-path mimicking} makes it easier to learn useful information from T-head. Furthermore, we propose the \textit{staggered feature pyramid} from which one can extract a pair of features with different resolutions. For each anchor box, it obtains high-resolution features for T-head and low-resolution features for S-head from different layers of the feature pyramid. Thus MimicDet can benefit more from high-accuracy of T-head and high-efficiency of S-head without extra computation cost.

To demonstrate the effectiveness of MimicDet, we conduct experiments on the challenging COCO detection benchmark and its performance significantly surpasses other state-of-the-art one-stage detectors. Our best model, based on ResNeXt-101 backbone, achieves state-of-the-art 46.1 mAP on COCO \textit{test-dev} benchmark.

\section{Related Work}
\label{related works}

In this section we briefly review related works in the fields of object detection and network mimicking.

\textbf{Two-stage detectors}: For now, two-stage detectors take the lead in detection accuracy. In these detectors, sparse region proposals are generated in the first stage and then are further regressed and classified in the second stage. R-CNN~\cite{girshick2014rich} utilized low-level computer vision algorithms such as Selective Search~\cite{uijlings2013selective} and Edge Boxes~\cite{zitnick2014edge} to generate proposals, then adopt CNN to extract features for training SVM classifier and bounding box regressor. SPP-Net~\cite{he2014spatial} and Fast R-CNN~\cite{girshick2015fast} proposed to extract features for each proposal on a shared feature map by spatial pyramid pooling. Faster R-CNN~\cite{ren2015faster} integrated the region proposal process into the deep convnet and makes the entire detector an end-to-end trainable model. R-FCN~\cite{dai2016r} proposed a region-based fully convolutional network to generate region-sensitive features for detection. FPN~\cite{lin2017feature} proposed a top-down architecture with lateral connection to generate a feature pyramid for multi-scale detection. In our work, instead of detecting object directly, the two-stage detection head performs as a teacher model in mimic mechanism and will not be used for inference.

\textbf{One-stage detectors}: One-stage detectors perform classification and regression on dense anchor boxes without generating a sparse RoI set. YOLO~\cite{redmon2016you} is an early exploration that directly detects objects on dense feature map.  SSD~\cite{liu2016ssd} proposed to use multi-scale features for detecting variant scale objects. RetinaNet~\cite{lin2017focal} proposed focal loss to address the extreme class imbalance problem in dense object detection. RefineDet~\cite{zhang2018single} introduced anchor refinement module and the object detection module to imitate the cascade regression on dense anchor boxes. Guided Anchor~\cite{wang2019region} first used anchor-guided deformable convolutions to align features for RPN. AlignDet~\cite{chen2019revisiting} designed RoI convolution, to imitate RoI Align operation in two-stage detectors and perform exact feature alignment for densely anchor box detection.

\textbf{Network mimicking}: Typical network mimicking or knowledge distillation approaches utilize output logits or intermediate features of a well-trained large teacher model to supervise the training process of the efficient student model~\cite{bucilua2006model,hinton2015distilling,romero2014fitnets,sergey2017attention}. These approaches are widely used for model acceleration and compression in the classification task. Recently, ~\cite{li2017mimicking,chen2017learning} extend mimicking approaches to the object detection task and achieve satisfactory results.

Although we utilize mimicking as the critical method to improve the performance, MimicDet differs from the previous mimic-based object detection approaches in many aspects. First, the MimicDet use mimic between different types of detection heads instead of backbones. The teacher and student in MimicDet share the same backbone. Second, only an additional head branch as a teacher is introduced instead of an entire teacher model. Moreover, different from the pipeline in previous methods that train the teacher first and then mimic, teacher head and student head in MimicDet are trained jointly, which makes the training process much more efficient. Last, in conventional methods, student model usually needs a larger model to mimic, which limits the method to be applied on larger student model. MimicDet still works well in this situation because it does not rely on a stronger backbone to improve performance.

\begin{figure*}[t]
\centering
\includegraphics[width=1\linewidth]{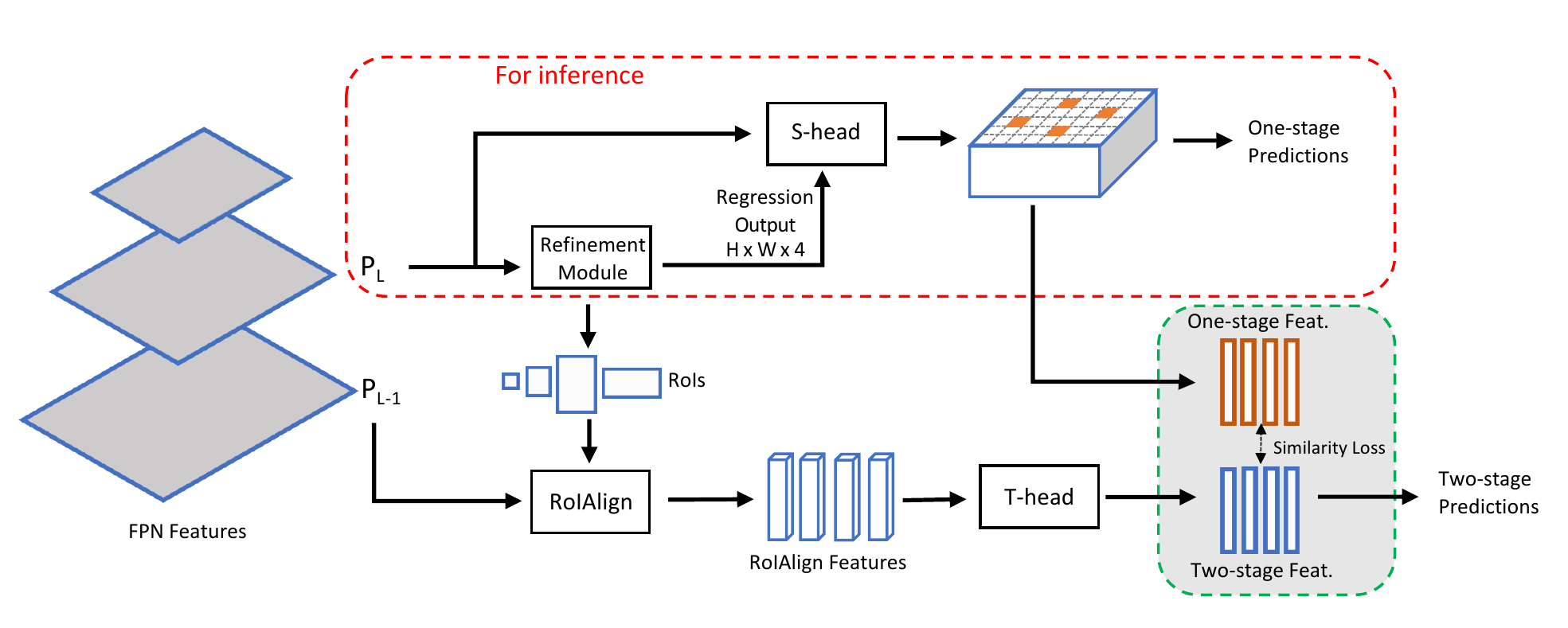} 
\caption{\textbf{Overview.} In training phase, we first adopt the refinement module to classify and regress the pre-defined dense anchor boxes. Then these adjusted anchor boxes will be transferred to S-head and T-head for dense and sparse detection, respectively. S-head takes the regression output of the refinement module in to compute the offset for deformable conv layer. Then it performs dense detection for each anchor box. T-head performs two-stage detection for sampled anchor boxes. Features of these sampled anchor boxes in S-head and T-head will be extracted for mimicking. During the inference, only the one-stage part(in red dash line) will be used.}
\label{fig:overview}
\end{figure*}

\section{MimicDet Framework}
\label{method}
The overview of the MimicDet is shown in Figure~\ref{fig:overview}. During the inference phase, MimicDet consists of backbone, refinement module, and one-stage detection head called S-head. In the training phase, we add an extra two-stage detection head as teacher, called T-Head, to guide the S-Head and enhance its performance. We will expatiate on each component in following sections.
\subsection{Backbone and Staggered Feature Pyramid}
We build our model on a Feature Pyramid Network~\cite{lin2017feature} backbone to efficiently extract multi-scale features for detecting objects that distributed in a wide range of scales. Generally, the FPN network adopts lateral connections to combine bottom-up feature with the top-down feature and generates a feature pyramid for further process. Here we follow the conventional denotation in the FPN, using $C_l$ and $P_l$ to denote the feature in ResNet~\cite{he2016deep} and FPN feature pyramid, respectively. Since the one-stage detection head is sensitive to the computation cost, we follow the modification in RetinaNet~\cite{lin2017focal} and extend the original feature pyramid to $P_7$, which has a stride of 128 with respect to the input image. Specifically, $P_6$ is generated by adopting a 3$\times$3 stride-2 convolution on $C_5$, $P_7$ is computed by applying ReLU~\cite{nair2010rectified} followed by a 3$\times$3 stride-2 conv on $P_6$. The number of channels for all feature pyramid is 256.

A subtle but crucial modification in backbone is that we keep the pyramid features from $P_2$ to $P_7$ in the training phase instead of $P_3$ to $P_7$. Then we build the staggered feature pyramid by grouping these features into two sets: $\{P_2, P_3, P_4, P_5, P_6\}$ and $\{P_3, P_4, P_5, P_6, P_7\}$. The former high-resolution feature set is used for the T-head, while the latter low-resolution one is used for the S-head as well as refinement module. This arrangement meets the different needs of the one-stage head and the two-stage head. As for one-stage heads in MimicDet, efficiency is the first priority, while the detection accuracy is more critical in second-stage head which performs as a teacher model in mimicking. Low-resolution features help the S-head to detect faster and high-resolution features make the T-head detect accurate. At inference phase, introducing staggered feature pyramid is cost-free because we only use the low-resolution one and do not generate the $P_2$ for efficiency.%

\subsection{Refinement Module}
After extracting features from the FPN, we adopt the refinement module to filter out easy negatives and adjust the location and size of pre-defined anchor boxes, which can mitigate the extreme class imbalance issue and provide better anchor initialization when training S-head and T-head. Specifically, the refinement module is formed by one 3$\times$3 conv and two sibling 1$\times$1 convs to perform class-agnostic binary classification and bounding box regression on top of the feature pyramid. Anchor boxes that adjusted by the refinement module will be transferred to T-head and S-head for sparse and dense detection, and only top-ranked boxes will contribute to the training process in T-head and S-head. In general, to MimicDet, the refinement module plays a similar role as the RPN in FPN~\cite{lin2017feature}(two-stage detector) and the ADM in RefineDet~\cite{zhang2018single}(one-stage detector).

We define anchors to have areas from $32^2$ to $512^2$ on feature pyramid $P_3$ to $P_7$, respectively. Unlike previous works which define multiple anchors on a certain position of the feature map, we only define one anchor on each position with 1:1 aspect ratio. We adopt this sparse anchor setting to avoid feature sharing because each anchor box in the one-stage head needs to have an exclusive and explicit feature for head mimicking. 

Assigning label of the anchors in the refinement module is different from the conventional RoI based strategy because the pre-defined anchor boxes are much more sparse than those in RetinaNet. We assign the objects to feature pyramid from $P_3$ to $P_7$ according to their scale, and each feature pyramid learns to detect objects in a particular range of scales. Specifically, for pyramid $P_l$, the valid scale range of the target object is computed as $[S_l\times\eta_1, S_l\times\eta_2]$, where $S_l$ is the basic scale for level $l$, and the $\eta$ is set to control the valid scale range. We empirically set $S_l = 4\times2^l$, $\eta_1=1, \eta_2=2$. Any objects that smaller than $S_3\times\eta_1$ or larger than $S_7\times\eta_2$ will be assigned to $P_3$ or $P_7$, respectively. Given a ground-truth box $B$, we define its positive area as its central rectangle area that shrunk to 0.3 times of the original length and width. Suppose there exists a ground-truth box $B$ lies in the valid scale range of an anchor box $A$, and the central point of $A$ lies in the positive area of $B$. Only in this case, the $A$ will be labeled as a positive sample, and the regression target is set to $B$. The regression target is kept the same as Faster R-CNN~\cite{ren2015faster}, and only positive anchors will contribute to the regression loss. We optimize the classification and regression branch by binary cross-entropy loss and $L1$ loss, respectively.

\begin{figure}[t]
\centering
\includegraphics[width=1\linewidth]{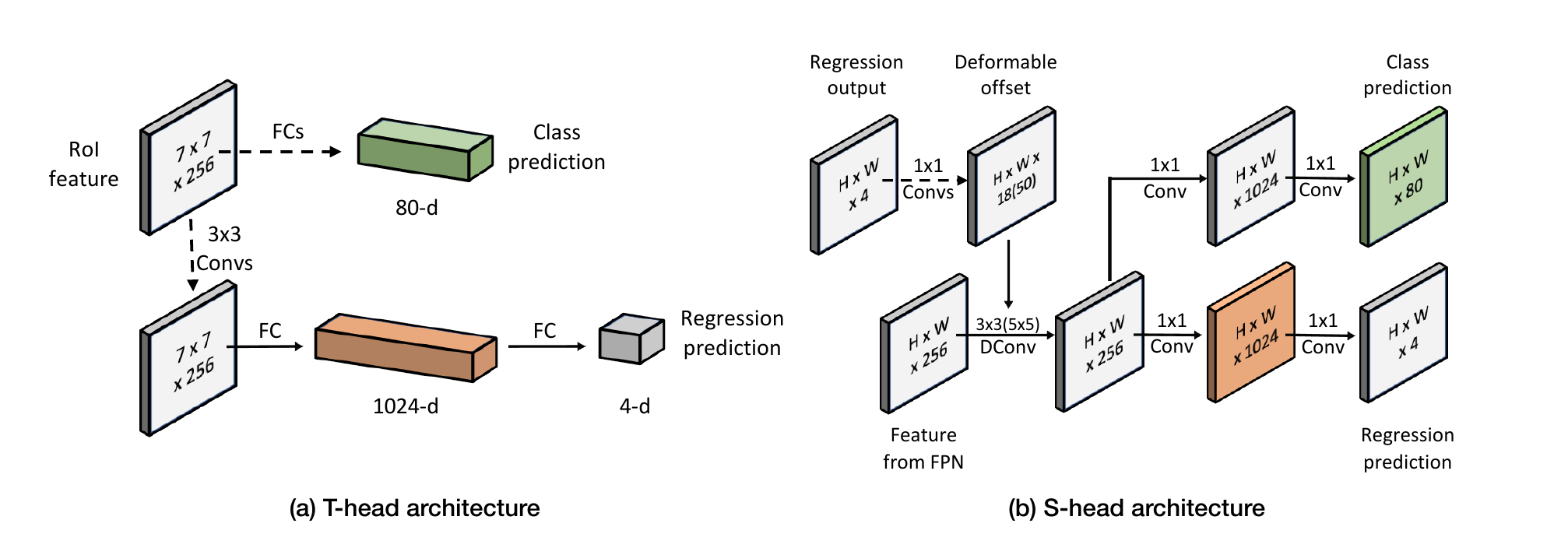} 
\caption{\textbf{Detection head architecture.} Dash line denotes stacked operations. Green and orange features will be used for dual-path head mimicking, i.e. classification and regression mimicking.}
\label{fig:architecture}
\end{figure}

\subsection{Detection Heads}
\textbf{T-head: }
T-head is a two-stage detection head that detects a sparse set of the anchor boxes adjusted by refinement module. The purpose of the T-head is to extract high-quality features and using these features to guide the training process of the S-head. We design the T-head as a heavy head with high resolution input for better features and the T-head will not be used during the inference.

As shown in Figure~\ref{fig:overview}, T-head access features from the high-resolution set of the staggered feature pyramid, i.e., one layer ahead of the original one. It first adopts the RoIAlign operation to generate location-sensitive features with $7\times7$ resolution for each anchor box. After that, the T-head will be split into two branches for classification and regression. In the classification branch, features of each anchor box are processed by two 1024-d fc layers, followed by one 81-d fc layer and a softmax layer to predict the classification probability. In the other branch, we adopt four consecutive $3\times3$ convs with 256 output channels and then flat the feature into a vector. Next, the flatten feature will be transformed into a 1024-d regression feature and a 4-d output sequentially by two fc layers. As shown in Figure~\ref{fig:architecture}, the 81-d classification logits and 1024-d regression feature will be regarded as the mimicking target when training the S-head. The label assignment is based on the IoU criteria with 0.6 threshold. We optimize the classification and regression branch by cross-entropy loss and $L1$ loss respectively.

\textbf{S-head:}
S-head is a one-stage detection head that directly performs dense detection on top of the feature pyramid without sampling. We design the S-head as a light-weight network that can overcome the feature misalignment and learning to extract high-quality features by mimicking the T-head. As we mentioned before, introducing the refinement module will break the location consistency between the anchor box and its corresponding features. The location inconsistency can lead to differences in representing area between S-head and T-head, which is harmful to head mimicking. So we use deformable convolution to capture the misaligned feature. The deformation offset is computed by a micro-network that takes the regression output of the refinement module as input.
The architecture of S-head is shown in Figure~\ref{fig:architecture}. The micro-network is formed by three 1$\times$1 convs with 64 and 128 intermediate channels. After that we use one 5$\times$5 deformable conv(with 256 channels) and two sibling 1$\times$1 convs to extract 1024-d features for classification and regression branch. Then two individual 1$\times$1 convs are used for generating predictions. To further reduce the computation, we replace the 5$\times$5 deformable conv with 3$\times$3 deformable conv in the highest resolution level $P_3$. The label assignment strategy and loss functions are kept same as those of the T-head for semantic consistency.

During the experiments we found that the proportion of positive samples of particular classes is too low, even though the refinement module has already rejected some easy negative samples. To deal with it, we use hard negative mining to mitigate the class imbalance problem, i.e., we always sample boxes with top classification loss to optimize the classification loss in S-head.

\subsection{Head Mimicking}
We use $B_s$ to denote the set of all anchor boxes adjusted by the refinement module, and $B_t$ to denote the sparse subset of the $B_s$ that sampled for T-head. Based on $B_t$, we define $B_m$ as a random sampled subset used for optimizing mimic loss. Given $B_m$, we can get its corresponding two-stage classification feature set $F^{tc}$ and regression feature set $F^{tr}$ by applying T-head on it. Similarly, classification and regression features of $B_m$ in S-head can also be obtained and denoted as $F^{sc}$ and $F^{sr}$. More specifically, in S-head, each pixel of its output feature map corresponds to an anchor box in $B_s$. To get the S-head feature of an adjusted anchor box, we trace back to its initial position and extract the pixel at that position in the feature map of S-head.

We define the mimic loss as follow:
\begin{equation}
\begin{aligned}
    L_{mimic} =&\frac{1}{N_{B_m}}\{\sum\nolimits_{i}[1-cosine(F^{tr}_i,F^{sr}_i)] \\ &+\sum\nolimits_{i}[1-cosine(F^{tc}_i,F^{sc}_i)]\}
\end{aligned}
\end{equation}
where $i$ is the index of anchor boxes in ${B_m}$ and $N_{B_m}$ is the number of the anchor boxes in ${B_m}$. We adopt cosine similarity to evaluate the similarity loss on both classification and regression branches.

Finally, we formally define the multi-task training loss as: $L=L_R+L_S+L_T+L_{mimic}$, where $L_R$,$L_S$ and $L_T$ denotes loss of the refinement module, S-head and T-head respectively.


\section{Implementation Details}
\subsection{Training}
We adopt the depth 50 and 101 ResNet with FPN as our default backbone to generate the feature pyramid. Images are resized such that their shorter side is 800 pixels by convention. When training the refinement module, different from RPN, we utilize all anchor boxes with no sampling strategy involved. After that, we adopt NMS with 0.8 IoU threshold on anchor boxes that adjusted by refinement module and select top 2000 boxes as the proposal set. In T-head, 128 boxes are sampled from the proposal set with 1:3 ratio of positive to negative samples. In S-head, we sort boxes in the proposal set by their classification losses and select 128 boxes with top loss value for training the classification branch. As for the regression branch, we sample at most 64 positive boxes from the proposal set randomly. In addition to that, all 128 boxes that are sampled for T-head will be adopted to optimize the mimic loss.

We use SGD to optimize the training loss with 0.9 momentum and 0.0001 weight decay. The backbone parameters are initialized with the ImageNet-1k based pre-trained classification model. New parameters in FPN and T-head are initialized with He initialization~\cite{he2015delving}, others newly added parameters are initialized by Gaussian initializer with $\sigma=0.01$. By convention, we freeze the parameters in the first two conv layers and all batchnorm layers. No data augmentations except standard horizontal flipping are used. We set two default learning schedules named 1$\times$ and 2$\times$, which has 13 and 24 epochs in total. The learning rate is set to 0.04 and then decreases by 10 at the 8th and 11th epoch in 1$\times$ schedule or 16th and 22nd epoch in 2$\times$ schedule. All models are trained on 16 GPUs with a total batch size of 32. We use warmup mechanisms in both baselines and our method to keep the multi-GPU training stable.
\subsection{Inference}
During inference, we discard the T-head and do not compute the $P_2$ feature map in FPN. The initial anchor boxes will be regressed twice by refinement module and S-head. To reduce computational consumption in post-processing, we filter out the top 1000 boxes according to their classification score predicted by refinement module. Then these boxes will be processed by NMS with 0.6 IoU threshold and 0.005 score threshold. Finally, the top 100 scoring boxes will be selected as the final detection result.

\section{Experiments}
\label{sec:exp}
We conduct experiments on the object detection track of the COCO benchmark and compare our method with other state-of-the-art detectors. All experiments below are trained on the union of 80k train images and 35k subset of val images, and the test is performed on the 5k subset of val (minival) and 20k test-dev.

\subsection{Ablation Study}
For simplicity, unless specified, all experiments in ablation study are conducted in 1$\times$ learning schedule, with 800 pixels input size and ResNet-50 backbone. Experiment results are evaluated on COCO \textit{minival} set.

\textbf{How important is the head mimicking?} To measure the impact of the head mimicking mechanism, here we design and train three models: (a) MimicDet without T-head and mimic loss; (b) MimicDet with T-head and without mimic loss; (c) MimicDet. Experiment results are listed in Table~\ref{tab:mimic_mechanism}. Model (a) only achieves 36.0 AP, which is close to the RetinaNet baseline(we do not use focal loss in MimicDet). Model (b) achieves 36.4 AP and gains 0.4 AP comparing with model (a). Model (c) gets the best result with 39.0 AP, which achieves 2.6 AP improvement over model (b). These results demonstrate that adding a two-stage head hardly affects the performance of one-stage head, and the mimic mechanism is the principal factor of MimicDet's success.

\begin{figure}[t!]
\small
\centering
\includegraphics[height=120pt]{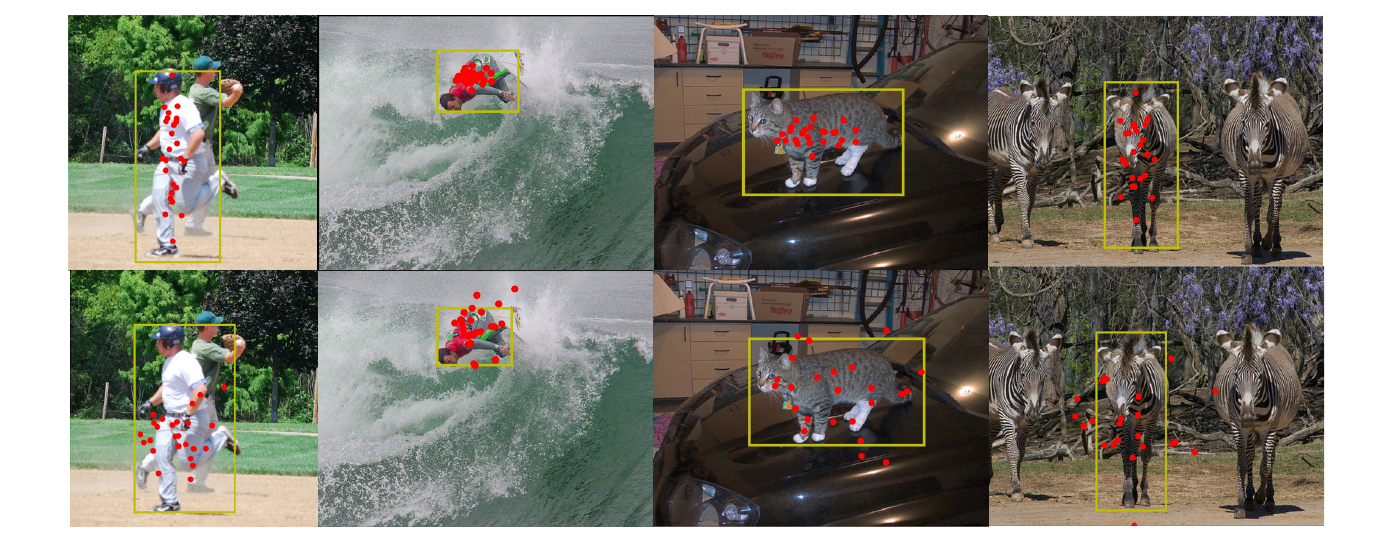}
\caption{\textbf{MimicDet(top) \textit{vs.} MimicDet w/o head-mimicking mechanism(bottom).} Yellow box is the adjusted anchor box generated by refinement module. It's corresponding deformable convolution locations are denoted by red dots.}
\label{fig:visualization}
\end{figure}

In Figure~\ref{fig:visualization}, we compare models that trained with and without mimic mechanism by visualizing the location consistency between their features and corresponding boxes. Although the plain guided deformable convolution has the potential for fixing the feature misalignment issue, results in practice are not that satisfactory. With the proposed mimic mechanism, guided deformable convolution learns to align the representing area between one-stage and two-stage branches, which helps to close the semantic gap between them. On the other hand, S-head can capture features precisely and consistently, which helps to mitigate the feature misalignment in one-stage detection.

\setlength{\tabcolsep}{5pt}
\begin{table}[h]
\small
\begin{center}
\caption{An ablation study to inspect the impact of introducing T-head and head-mimicking mechanism.}
\begin{tabular}{ l  c  c  c  }
\hline
\noalign{\smallskip}
Method & AP & $\text{AP}_{.5}$ & $\text{AP}_{.75}$  \\
\noalign{\smallskip}
\hline
\noalign{\smallskip}
w/o T-head \& w/o mimic & 36.0 & 56.0 & 39.9 \\
\noalign{\smallskip}
w T-head \& w/o mimic & 36.4 & 56.1 & 40.0  \\
\noalign{\smallskip}
w T-head \& w mimic & 39.0 & 57.5 & 42.6 \\
\hline
\end{tabular}
\label{tab:mimic_mechanism}
\end{center}
\end{table}

\textbf{Design of S-head:} In this section we explore the variant design of S-head. As discussed before, S-head should be designed as light as possible to ensure the computation efficiency. To this end, we first fix the architecture of S-head as a three-layer convolutional neural network. The first layer, called alignment layer, is adopted to capture the misaligned feature caused by cascade regression. It helps to align the representing area of S-head and T-head and close the semantic gap between them. The second convolutional layer project the aligned feature to a high dimensional feature space, and the third layer performs the final prediction. For simplicity, we keep second and third layer as 1$\times$1 conv layer unchanged and explore the design of the alignment layer.

As shown in Table~\ref{tab:variant_s_head}, we experiment on four typical variants denoted as (a) to (d). For (a), the alignment layer is a normal conv layer; For (b), it is a deformable conv layer; For (c), the original deformable conv is replaced by guided deformable conv layer. More specifically, the offsets in deformable conv layer is computed by the regression output of the refinement module. For (d), we replace 5$\times$5 guided deformable conv with the 3$\times$3 one for S-head on $P_3$ to reduce the computation cost. 

Experiments in Table~\ref{tab:variant_s_head} demonstrate that the alignment layer of S-head has marked impact on mimicking result. Without an efficient alignment layer, the representing areas in S-head and T-head will be different, and the performance of our mimic workflow deteriorates severely because of the large semantic gap between them. Model (b) and (c) yields 1.5 and 5.5 AP improvement over the model (a) respectively. Model (d) achieves 39.0 AP, which is comparable with model (c). For better speed-accuracy trade-off, we choose (d) as the default setting in MimicDet.

\setlength{\tabcolsep}{5pt}
\begin{table}[h]
\small
\begin{center}
\caption{Comparison of S-head variants with different alignment layers. $^*$ indicates using $3\times3$ guided deformable conv for the S-head on $P_3$ layer.}

\begin{tabular}{ l  c  c  c  }
\hline
\noalign{\smallskip}
 Method & AP & $\text{AP}_{.5}$ & $\text{AP}_{.75}$  \\
 \noalign{\smallskip}
\hline
\noalign{\smallskip}
(a) $5\times5$ normal conv & 33.7 & 54.1 & 37.2 \\
\noalign{\smallskip}
(b) $5\times5$ deform conv & 35.2 & 55.3 & 39.2  \\
\noalign{\smallskip}
(c) $5\times5$ guided deform conv & 39.2 & 57.3 & 43.1 \\
\noalign{\smallskip}
(d) $5\times5$ guided deform conv$^*$ & 39.0 & 57.5 & 42.6 \\
\hline
\end{tabular}
\label{tab:variant_s_head}
\end{center}
\end{table}

\textbf{Design of T-head: }In this section we inspect how the design of T-head impacts the mimicking result in MimicDet. We adopt a light-weight T-head as our baseline. In this baseline, T-head is formed by two consecutive 1024-d fully connected layers and two sibling output layers, which is a common baseline choice in many two-stage detectors. To further enhance the performance of T-head, we try to introduce a heavier architecture with conv layers. The heavy head consists of four conv layers followed by two 1024-d fully connected layers and two sibling output layers. Note that in all T-head variants above, features in both S-head and T-head are not split and the mimicking is performed on the last 1024-d vector features.

From the second row of Table~\ref{tab:variant_t_head}, we observe that the heavy design does increase the AP of T-head as expected, but the AP of S-head deteriorates instead. This observation suggests that, as the complexity of T-head increases, the difficulty in mimicking also boosts simultaneously. In MimicDet, we propose dual-path mimicking to decompose the feature of classification and regression into two branches and conduct two mimicking processes individually. Comparing the second row and the third row of Table~\ref{tab:variant_t_head}, the proposed dual-path mimicking achieves 39.0 AP in S-head, with a 1.9 AP gain from the naive heavy-head baseline.

\setlength{\tabcolsep}{5pt}
\begin{table}[h]
\small
\begin{center}
\caption{Comparison of T-head variants. The evaluation settings of T-head are same as those in S-head. Arrows denote the deterioration or amelioration in comparison with light head design.}
\begin{tabular}{ l  c  c  }
\hline\noalign{\smallskip}
Method & T-head AP &  S-head AP \\
\noalign{\smallskip}
\hline
\noalign{\smallskip}
Light head & 39.0 & 37.5 \\
\noalign{\smallskip}
Heavy head & 39.8 $\uparrow$ & 37.1 $\downarrow$ \\
\noalign{\smallskip}
Dual-path mimicking & 40.2 $\uparrow$ & 39.0 $\uparrow$ \\
\hline
\end{tabular}
\label{tab:variant_t_head}
\end{center}
\end{table}

\textbf{Staggered feature pyramid:} In MimicDet we propose staggered feature pyramid to obtain features for T-head and S-head in different layers of the feature pyramid. In first row of Table~\ref{tab:staggered}, we use feature pyramid $\{P_3-P_7\}$ for both S-head and T-head. In the second row, we use high-resolution feature pyramid $\{P_2-P_6\}$ for T-head and low-resolution feature pyramid $\{P_3-P_7\}$ for S-head. Results in Table~\ref{tab:staggered} show that, by mimicking high-resolution features, MimicDet gains 0.6 AP improvement in ResNet-50 experiment.

\setlength{\tabcolsep}{5pt}
\begin{table}[h]
\small
\begin{center}
\caption{Ablation study for staggered feature pyramids. The second and third column denotes the layers in feature pyramid that will be used for T-head and S-head, respectively.}
\begin{tabular}{ l  c  c  c  }
\hline\noalign{\smallskip}
Method & T-head  & S-head & AP  \\
\noalign{\smallskip}
\hline
\noalign{\smallskip}
RetinaNet FPN~\cite{lin2017focal} & $P_3$ - $P_7$ & $P_3$ - $P_7$ & 38.4 \\
\noalign{\smallskip}
staggered feature pyramid & $P_2$ - $P_6$ & $P_3$ - $P_7$ & 39.0 \\
\hline
\end{tabular}
\label{tab:staggered}
\end{center}
\end{table}

\textbf{Shared backbone versus separated backbone:} The teacher and student in MimicDet share a same backbone. Here we compare MimicDet with its separated backbone counterpart. In separated backbone experiments, we will first train a two-stage detector as the teacher, then the pre-trained teacher model will be adopted to supervise the one-stage student model. Note that we inherit all the design in MimicDet except sharing the backbone. We conduct experiments on 16 Titan Xp GPUs. Results in Table~\ref{tab:split_backbone} demonstrate that MimicDet achieves comparable accuracy with only about half of the training time in comparison with separated backbone counterpart.

\setlength{\tabcolsep}{8pt}
\begin{table}[h]
\small
\begin{center}
\caption{Training time in comparison with separated backbone counterpart. The learning rate schedule is set to 2x.}
\begin{tabular}{c  c  c  c }
\hline
\noalign{\smallskip}
Backbone &  & AP & training time  \\
\noalign{\smallskip}
\hline
\noalign{\smallskip}
Res-50 & shared & 39.8 & 17.5h \\
\noalign{\smallskip}
Res-50 & separated & 40.2 & 34h \\
\noalign{\smallskip}
Res-101 & shared & 41.8 & 22h \\
\noalign{\smallskip}
Res-101 & separated & 42.2 & 47h \\
\hline
\end{tabular}
\label{tab:split_backbone}
\end{center}
\end{table}

\textbf{Speed/Accuracy trade-off:} Larger input size and backbone networks yield better accuracy, but also slower inference speed. we conduct experiments on variant input sizes and variant backbone networks to show the speed-accuracy trade-off, results are listed in Table~\ref{tab:variant_input_size}. On single TITAN V GPU, MimicDet achieves 23 FPS with 38.7 mAP on Res-50 with 600 pixels input size and 13 FPS with 41.8 mAP on Res-101 with 800 pixels input size. 

\setlength{\tabcolsep}{6pt}
\begin{table}[h]
\scriptsize
\begin{center}
\caption{The speed-accuracy trade-off by varying image sizes. We test the inference speed on single TITAN V GPU with batch size of 1. The learning rate schedule is set to 2$\times$.}
\begin{tabular}{c c c c c c c c c}
\hline\noalign{\smallskip}
Size & Backbone & AP & $\text{AP}_{.5}$ & $\text{AP}_{.75}$ & $\text{AP}_{S}$ & $\text{AP}_{M}$ & $\text{AP}_{L}$ & Inference time\\
\noalign{\smallskip}
\hline
\noalign{\smallskip}
 400 & ResNet-50 & 35.5 & 53.1 & 38.2 & 15.3 & 38.4 & 52.3 & 34ms \\
 \noalign{\smallskip}
 600 & ResNet-50 & 38.7 & 56.8 & 42.2 & 20.3 & 41.7 & 54.5 & 44ms \\
 \noalign{\smallskip}
 800 & ResNet-50 & 39.8 & 58.2 & 43.6 & 21.6 & 43.0 & 53.4 & 63ms \\
 \noalign{\smallskip}
 400 & ResNet-101 & 36.1 & 53.5 & 39.1 & 14.9 & 39.2 & 53.8 & 38ms \\
 \noalign{\smallskip}
 600 & ResNet-101 & 39.8 & 57.9 & 43.3 & 20.1 & 43.2 & 55.8 & 55ms \\
 \noalign{\smallskip}
 800 & ResNet-101 & 41.8 & 60.3 & 45.6 & 22.5 & 45.2 & 56.4 & 80ms \\
\hline
\end{tabular}
\label{tab:variant_input_size}
\end{center}
\end{table}

\subsection{Comparison with State-of-the-art Methods}
We compare MimicDet to other state-of-the-art one-stage object detectors on COCO test-dev. We adopt ResNet-101 and ResNeXt-101-64$\times$4d as our backbones. Follow the convention in RetinaNet, MimicDet is trained with scale jitter from 640 to 800 and 1.5$\times$ longer than the 2$\times$ training schedule.  As shown in Table~\ref{tab:test-dev}, MimicDet based on ResNet-101 and ResNeXt-101-64x4d achieves 44.4 AP and 46.1 AP respectively, which surpasses the current state-of-the-art one-stage methods by a large margin.

\setlength{\tabcolsep}{1pt}
\begin{table*}[h]
\scriptsize
\begin{center}
\caption{Comparison with state-of-the-art one stage detectors on COCO \textit{test-dev}. M means M40 or TitanX MaxWell, P means P100, Titan X(Pascal), Titan Xp or 1080Ti, V means TitanV. * indicates the results are based on flip test and soft NMS.}
\begin{tabular}{ c  c  c  c  c  c  c  c  c  c }
\hline
\noalign{\smallskip}
Method & Backbone & Input Size & Speed(ms) & AP & $\text{AP}_{.5}$ & $\text{AP}_{.75}$ & $\text{AP}_{S}$ & $\text{AP}_{M}$ & $\text{AP}_{L}$ \\
\noalign{\smallskip}
\hline
\noalign{\smallskip}
SSD~\cite{liu2016ssd} & ResNet-101 & 513 & 125/M & 31.2 & 50.4 & 33.3 & 10.2 & 34.5 & 49.8 \\
\noalign{\smallskip}
DSSD~\cite{fu2017dssd} & ResNet-101 & 513 & 156/M & 33.2 & 53.3 & 35.2 & 13.0 & 35.4 & 51.1 \\
\noalign{\smallskip}
RefineDet~\cite{zhang2018single} & ResNet101 & 512 & -  & 36.4 & 57.5 & 39.5 & 16.6 & 39.9 & 51.4 \\
\noalign{\smallskip}
YOLOv3~\cite{redmon2018yolov3} & DarkNet-53 & 608 & 51/M & 33.0 & 57.9 & 34.4 & 18.3 & 35.4 & 41.9 \\
\noalign{\smallskip}
ExtremeNet$^*$~\cite{zhou2019bottom} & Hourglass-104 & 511 & 322/P & 40.1 & 55.3 & 43.2 & 20.3 & 43.2 & 53.1 \\
\noalign{\smallskip}
CornerNet$^*$~\cite{law2018cornernet} & Hourglass-104 & 511 & 300/P & 40.5 & 56.5 & 43.1 & 19.4 & 42.7 & 53.9 \\
\noalign{\smallskip}
CenterNet$^*$~\cite{duan2019centernet} & Hourglass-104 & 511 & 340/P & 44.9 & 62.4 & 48.1 & 25.6 & 47.4 & 57.4 \\
\noalign{\smallskip}
\hline
\noalign{\smallskip}
RetinaNet~\cite{lin2017focal} & ResNet-101 & 800 & 104/P 91/V & 39.1 & 59.1 & 42.3 & 21.8 & 42.7 & 50.2 \\
\noalign{\smallskip}
FSAF~\cite{zhu2019feature} & ResNet-101 & 800 & 180/M & 40.9 & 61.5 & 44.0 & 24.0 & 44.2 & 51.3 \\
\noalign{\smallskip}
FCOS~\cite{tian2019fcos} & ResNet-101 & 800 & 86.2/V & 41.5 & 60.7 & 45.0 & 24.4 & 44.1 & 51.0 \\
\noalign{\smallskip}
RPDet~\cite{yang2019reppoints} & ResNet-101 & 800 & - & 41.0 & 62.9 & 44.3 & 23.6 & 44.1 & 51.7 \\
\noalign{\smallskip}
AlignDet~\cite{chen2019revisiting} & ResNet-101 & 800 & 110/P & 42.0 & 62.4 & 46.5 & 24.6 & 44.8 & 53.3 \\
\noalign{\smallskip}
\textbf{MimicDet(ours)} & ResNet-101 & 800 & 80/V &  \textbf{44.4} & 63.1 & 48.8 & 25.8 & 47.7 & 55.8 \\
\noalign{\smallskip}
\hline
\noalign{\smallskip}
RetinaNet~\cite{lin2017focal} & ResNeXt-101-32$\times$8d & 800 & 177/P & 40.8 & 61.1 & 44.1 & 24.1 & 44.2 & 51.2 \\
\noalign{\smallskip}
FSAF~\cite{zhu2019feature} & ResNeXt-101-64$\times$4d & 800 & 362/M & 42.9 & 63.8 & 46.3 & 26.6 & 46.2 & 52.7 \\
\noalign{\smallskip}
FCOS~\cite{tian2019fcos} & ResNeXt-101-64$\times$4d & 800 & 143/V & 43.2 & 62.8 & 46.6 & 26.5 & 46.2 & 53.3 \\
\noalign{\smallskip}
\textbf{MimicDet(ours)} & ResNeXt-101-64$\times$4d & 800 & 132/V & \textbf{46.1} & 65.1 & 50.8 & 28.6 & 49.2 & 57.5 \\
\noalign{\smallskip}
\hline
\end{tabular}
\label{tab:test-dev}
\end{center}
\end{table*}
\section{Conclusion}
In this paper, we propose a novel and efficient framework to train a one-stage detector called MimicDet. By introducing two-stage head with mimicking mechanism, MimicDet can obtain excellent properties of two-stage features and get rid of common issues that lie in the one-stage detectors. MimicDet inherits both high-efficiency of the one-stage approach and high-accuracy of the two-stage approach. It does not need to pre-train a teacher model or require a stronger backbone model to serve as the teacher. These properties make MimicDet much more efficient and can be easily extended to larger models in comparison with conventional mimic methods.


\clearpage
%
%
\bibliographystyle{splncs04}
\bibliography{egbib}
\end{document}